\documentclass[11pt,letterpaper]{article}
\usepackage{emnlp2016}
\usepackage{times}
\usepackage{latexsym}
\usepackage{natbib}
\usepackage{verbatim}
\usepackage{amsmath}
\usepackage{cases}
\usepackage{xcolor}
\usepackage{cleveref}
\usepackage{cancel}
\usepackage{url}

\emnlpfinalcopy

\title{Matrix Factorization using Window Sampling and Negative Sampling\\ for Improved Word Representations\Thanks{This is a preprint of the paper that will be presented at the 54th Annual Meeting of the Association for Computational Linguistics.}}

\author{Alexandre Salle\textsuperscript{1} \quad Marco Idiart\textsuperscript{2} \quad Aline Villavicencio\textsuperscript{1} \\
  \textsuperscript{1} Institute of Informatics \\
  \textsuperscript{2} Physics Department \\
  Universidade Federal do Rio Grande do Sul \\
  Porto Alegre, Brazil \\
  {\tt \{atsalle,avillavicencio\}@inf.ufrgs.br, idiart@if.ufrgs.br }}

\date{}

\begin{document}

\maketitle

\begin{abstract}
In this paper, we propose LexVec, a new method for generating distributed word representations that uses low-rank, weighted factorization of the Positive Point-wise Mutual Information matrix via stochastic gradient descent, employing
a weighting scheme that assigns heavier penalties for errors on frequent co-occurrences while still accounting for negative co-occurrence. Evaluation on word similarity and analogy tasks shows that LexVec matches and often outperforms state-of-the-art methods on many of these tasks.

\end{abstract}

\section{Introduction}
Distributed word representations, 
or word embeddings,  
have been successfully used in many NLP applications \citep{Turian2010word,collobert2011natural,socher2013parsing}.
Traditionally, word representations have been obtained using \emph{count-based} methods \citep{Baroni2014}, where the  co-occurrence matrix is derived directly from corpus counts \citep{Lin1998}   
or using association measures like Point-wise Mutual Information (PMI) \citep{Church1990} and Positive PMI (PPMI) \citep{Bullinaria2007,Levy2014linguisticregexplicit}.

Techniques for generating lower-rank representations  
have also been employed, such as PPMI-SVD \citep{Levy2015improvingdist} and GloVe \citep{Pennington2014}, both achieving state-of-the-art performance on a variety of tasks. 

Alternatively, vector-space models can be generated with    
\emph{predictive} methods, which generally outperform the count-based methods \citep{Baroni2014}, the most notable of which is Skip-gram with Negative Sampling (SGNS, \cite{Mikolov2013negative}), which uses a neural network to generate embeddings. It implicitly factorizes a shifted PMI matrix, and its performance has been linked to the weighting of positive and negative co-occurrences \citep{Levy2014implicitmatrix}.  

In this paper, we present Lexical Vectors (LexVec), a method for factorizing PPMI matrices that combines characteristics of all these methods. On the one hand, it uses  SGNS window sampling, negative sampling,   
and stochastic gradient descent (SGD) to minimize a loss function that weights frequent co-occurrences heavily but also takes into account negative co-occurrence. However, since PPMI generally outperforms PMI on semantic similarity tasks \citep{Bullinaria2007}, rather than implicitly factorize a shifted PMI matrix (like SGNS), LexVec explicitly factorizes the PPMI matrix.

This paper is organized as follows: First, we describe PPMI-SVD, GloVe, and SGNS ($\S$\ref{sec:related}) before introducing the proposed method, LexVec ($\S$\ref{sec:LexVec}), and evaluating it on word similarity and analogy tasks ($\S$\ref{sec:materials}). 
We conclude with an analysis of results and discussion of future work.

We provide source code for the model at \url{https://github.com/alexandres/lexvec}.

\section{Related Work}
\label{sec:related}
\subsection{PPMI-SVD}
\label{sec:ppmi}

Given a word \emph{w} and a symmetric window of \emph{win} context words to the left and \emph{win} to the right, the co-occurrence matrix of elements $M_{wc}$  is defined as the number of times a target word \emph{w} and the context word \emph{c} co-occurred in the corpus within the window.
The PMI matrix is defined as  
\begin{equation}
PMI_{wc} = \log \frac{M_{wc} \; M_{**}}{ M_{w*} \; M_{*c} }
\end{equation}
where '*' represents the summation of the corresponding index.
As this matrix is unbounded in the inferior limit, in most applications it is replaced by its positive definite version, PPMI, where negative values are set to zero.
The performance of the PPMI matrix on word similarity tasks can be further improved by using \emph{context-distribution smoothing} \citep{Levy2015improvingdist} 
and \emph{subsampling} the corpus \citep{Mikolov2013negative}. 
As word embeddings with lower dimensionality may improve 
efficiency and generalization \citep{Levy2015improvingdist}, the improved PPMI$^*$ matrix can be factorized as a product of two lower rank matrices.
\begin{equation}
PPMI^*_{wc} \simeq W_w  \tilde{W}_c^\top
\end{equation}
where $W_w$ and $\tilde{W}_c$ are d-dimensional row vectors corresponding to vector embeddings for the target and context words.
Using the truncated SVD of size $d$ yields the factorization $U \Sigma T^\top$ with the lowest possible $L_2$ error \citep{eckert36approximation}. 

\citet{Levy2015improvingdist} recommend using $W = U \Sigma^p \; $ as the word representations, as suggested by \citet{bullinaria2012extracting}, who borrowed the idea of weighting singular values from the work of \citet{caron2001experiments} on Latent Semantic Analysis. Although the optimal value of $p$ is highly task-dependent \citep{osterlund-odling-sahlgren:2015:EMNLP}, we set $p = 0.5$ as it has been shown to perform well on the word similarity and analogy tasks we use in our experiments \citep{Levy2015improvingdist}.  

\subsection{GloVe}
GloVe \citep{Pennington2014} factors the logarithm of the co-occurrence matrix $\hat{M}$ that considers the position of the context words in the window.
The loss function for factorization is
\begin{equation}
\label{eq:glove}
L_{wc}^{GloVe} = \frac{1}{2} f(\hat{M}_{wc})(W_w\tilde{W}^\top_{c} + b_w + \tilde{b}_c - \log \hat{M}_{wc})^2
\end{equation}
where $b_w$ and $\tilde{b}_c$ are bias terms, and $f$ is a weighting function defined as
\begin{equation}
 f(x) = \begin{cases}
        (x / x_{max})^\beta & if \; x < x_{max}\\
        1 & otherwise
        \end{cases}
\end{equation}
$W$ and $\tilde{W}$ are obtained by iterating over all non-zero 
$(w,c)$ cells in the co-occurrence matrix and minimizing eq. \eqref
{eq:glove} through SGD.

The weighting function (in eq. \eqref{eq:glove}) penalizes more heavily reconstruction error of frequent co-occurrences, improving on PPMI-SVD's $L_2$ loss, which weights all reconstruction errors equally. 
However, as it does not 
penalize reconstruction errors for pairs with zero counts in the co-occurrence matrix, no effort is made to scatter the vectors for these pairs.

\subsection{Skip-gram with Negative Sampling (SGNS)}
\label{sec:sgns}
SGNS \citep{Mikolov2013negative} 
trains a neural network to predict the probability of observing a context word $c$ given a target word $w$, sliding a symmetric window over a subsampled training corpus
with the window size being sampled uniformly from the range $[1, win]$. 
Each observed $(w,c)$ pair is combined with $k$ randomly sampled noise pairs $(w,w_i)$ and used to calculate the loss function
\begin{align}
\begin{split}
L_{wc}^{SGNS} &= \log \sigma(W_w\tilde{W_c}^\top) + \\ 
& \quad \sum\limits_{i=1}^k{\mathbf{E}_{w_i \sim P_n(w)} \log \sigma(-W_w\tilde{W}_{w_i}^\top)}
\end{split}
\end{align}
where $P_n(w)$ is the distribution from which noise words $w_i$  are sampled.\footnote{Following  \cite{Mikolov2013negative} it is the unigram distribution raised to the $3/4$ power.}
We refer to this routine which SGNS uses for selecting
$(w,c)$ pairs  by sliding a context window over the corpus for loss calculation and SGD as \emph{window sampling}.

SGNS is implicitly performing the weighted factorization of a shifted PMI matrix \citep{Levy2014implicitmatrix}. Window sampling ensures the factorization 
weights frequent co-occurrences heavily, but also takes into account negative co-occurrences, thanks to negative sampling.

\section{LexVec}
\label{sec:LexVec}
LexVec is based on the idea of factorizing the PPMI matrix  
using a reconstruction loss function that does not weight all errors equally, unlike SVD,
but instead penalizes errors of frequent co-occurrences more heavily, while still 
treating negative co-occurrences, unlike GloVe.  Moreover, given that using PPMI results in better performance than PMI on semantic tasks, we propose keeping the SGNS weighting scheme by using window sampling and negative sampling, but explicitly factorizing the PPMI matrix rather than implicitly factorizing the shifted PMI matrix. 
The LexVec loss function has two
terms 

\begin{footnotesize}
\begin{align}
\label{eq:lexvec2}
L_{wc}^{LexVec} &= \frac{1}{2} (W_w\tilde{W_c}^\top - PPMI_{wc}^*)^2 \\
\label{eq:lexvec3}
L_{w}^{LexVec} &= \frac{1}{2} \sum\limits_{i=1}^k{\mathbf{E}_{w_i \sim P_n(w)} (W_w\tilde{W_{w_i}}^\top - PPMI_{ww_i}^*)^2 }
\end{align}
\end{footnotesize}
We minimize \cref{eq:lexvec2,eq:lexvec3} using two alternative approaches:

\noindent\textbf{Mini-Batch (MB):}  
This variant executes gradient descent in exactly the same way as SGNS. 
Every time a pair $(w,c)$ is observed by window sampling and pairs $(w, w_{1...k})$ drawn by negative sampling, $W_w$, $\tilde{W}_c$, and $\tilde{W}_{w_{1...k}}$ are updated by gradient descent on the sum of eq.\eqref{eq:lexvec2} and eq.\eqref{eq:lexvec3}. 
The global loss for this approach is
\begin{equation}
\label{eq:lexvecmb}
L^{LexVec} = \sum\limits_{(w,c)} \#(w,c) \;  ( L_{wc}^{LexVec} + L_{w}^{LexVec} )
\end{equation}
where $\#(w,c)$ is the number of times $(w,c)$ is observed in the subsampled corpus. 

\noindent\textbf{Stochastic (St):} 
Every context window is extended with $k$ negative samples $w_{1...k}$. Iterative gradient descent of \cref{eq:lexvec2} is then run on pairs $(w,c_j)$, for $j=1,..,2*$\emph{win} and $(w,c_i)$, $j=1,..,k$ for each window.
The global loss for this approach is
\begin{align}
\begin{split}
\label{eq:lexvecstochasticglobal}
L^{LexVec'} &= \sum\limits_{(w,c)} \#(w,c) L_{wc}^{LexVec} + \\ & \sum\limits_{w} \#(w) L_{w}^{LexVec}
\end{split}
\end{align}
where $\#(w)$ is the number of times $w$ is observed in the subsampled corpus.

If a pair $(w,c)$ co-occurs frequently, $\#(w,c)$ will weigh heavily in both \cref{eq:lexvecmb,eq:lexvecstochasticglobal}, giving the desired weighting for frequent co-occurrences. 
The noise term, on the other hand, has corrections proportional to 
$\#(w)$ and $\#(w_i)$, for each pair $(w,w_i)$. It produces corrections in pairs that due to frequency should be in the corpus but are not observed, therefore accounting automatically for negative co-occurrences.

\begin{table*}[t]
  \centering
  \begin{footnotesize}
  \begin{tabular}{c|ccccccccc}
\hline
Method & WSim & WRel & MEN & MTurk & RW & SimLex-999 & MC & RG & SCWS \\
\hline
PPMI-SVD	&	.731	& .617	& .731	& .627	& .427	& .303	& .770	& .756	& .615 \\
GloVe	&	.719 & .607	& .736	& .643	& .400	& .338	& .725	& .774	& .573 \\
SGNS	&	.770	& .670	& \textbf{.763}	& \textbf{.675}	& .465	& \textbf{.339}	& .823	& .793	& \textbf{.643} \\
LexVec + MB + $WS_{PPMI}$ + $(W+\tilde{W})$	&	.770	& .671	& .755	& .650	& .455 & .322	& .824	& .830	& .623 \\
LexVec + St. + $WS_{PPMI}$ + $(W + \tilde{W})$	&	.763	& .671	& .760	& .655	& .458	& .336	& .816	& .827	& .630 \\
LexVec + MB + $WS_{PPMI}$ + $W$	&	.748	& .635	& .741	& .636	& .456	&	.320	& .827	&	.820	& .632 \\
LexVec + St. + $WS_{PPMI}$ + $W$	&	.741	& .622	& .733	& .628	& .457	&	.338	& .820	&	.808	& .638 \\
LexVec + MB + $WS_{SGNS}$ + $(W + \tilde{W})$	&	.768	& \textbf{.675}	& .755	& .654	& .448	&	.312	& .824	&	.827	& .626 \\
LexVec + St. + $WS_{SGNS}$ + $(W + \tilde{W})$	&	\textbf{.775}	& .673	& .762	& .654	& \textbf{.468}	&	\textbf{.339}	& \textbf{.838}	&	\textbf{.848}	& .628 \\
LexVec + MB + $WS_{SGNS}$ + $W$	&	.745	& .640	& .734	& .645	& .447	&	.311	& .814	&	.802	& .624 \\
LexVec + St. + $WS_{SGNS}$ + $W$	&	.740	& .628	& .728	& .640	& .459	&	\textbf{.339}	& .821	&	.818	& .638 \\
\hline
  \end{tabular}
  \end{footnotesize}
  \caption{Spearman rank correlation on word similarity tasks.}
  \label{tab:wordsim}
\end{table*}

\begin{table*}[t]
  \centering
  \begin{footnotesize}
  \begin{tabular}{c|ccc}
\hline
Method & \begin{tabular}[x]{@{}c@{}}GSem\\3CosAdd / 3CosMul\end{tabular} & \begin{tabular}[x]{@{}c@{}}GSyn\\3CosAdd / 3CosMul\end{tabular} & \begin{tabular}[x]{@{}c@{}}MSR\\3CosAdd / 3CosMul\end{tabular} \\
\hline
PPMI-SVD	& .460 / .498 &  .445 / .455	&  .303 / .313 \\
GloVe &	\textbf{.818} / .813	& .630 / .626	& .539 / \textbf{.547} \\
SGNS	&	.773 / .777	& .642 / \textbf{.644}	& .481 / .505 \\
LexVec + MB + $WS_{PPMI}$ + $(W+\tilde{W})$	&	.775 / .792	& .520 / .539	& .371 / .413 \\
LexVec + St + $WS_{PPMI}$ + $(W+\tilde{W})$ &	.794 / .807	&	.543 / .555 & .378 / .408 \\
LexVec + MB + $WS_{PPMI}$ + $W$ & .800 / .805	&	.584 / .597 & .421 / .457 \\
LexVec + St. + $WS_{PPMI}$ + $W$ & .787 / .782	&	.597 / .613 & .445 / .475 \\
LexVec + MB + $WS_{SGNS}$ + $(W + \tilde{W})$ & .762 / .785	&	.520 / .534 & .349 / .386 \\
LexVec + St. + $WS_{SGNS}$ + $(W + \tilde{W})$ & .792 / .809	&	.536 / .553 & .362 / .396 \\
LexVec + MB + $WS_{SGNS}$ + $W$ & .798 / .807	&	.573 / .580 & .399 / .435 \\
LexVec + St. + $WS_{SGNS}$ + $W$ & .779 / .778	&	.600 / .614 & .434 / .463 \\
\hline
  \end{tabular}
  \end{footnotesize}
  \caption{Results on word analogy tasks, given as percent accuracy.}
  \label{tab:analogies}
\end{table*}

\section{Materials}
\label{sec:materials}
All models were trained on a dump of Wikipedia from June 2015, split into sentences, with punctuation removed, numbers converted to words, and lower-cased. Words with less than 100 counts were removed, resulting in a vocabulary of 302,203 words. All models generate embeddings of 300 dimensions.

The PPMI* matrix used by both PPMI-SVD and LexVec was constructed using smoothing of $\alpha=3/4$ suggested in \citep{Levy2015improvingdist}  and an unweighted window of size 2. 
A dirty subsampling of the corpus is adopted for PPMI* 
and SGNS with threshold of $t=10^{-5}$ \citep{Mikolov2013negative}.\footnote{Words with unigram relative frequency $f>t$ are discarded from the training corpus with probability $p_w = 1 - \sqrt[]{t/f }$.} 
Additionally, SGNS uses $5$ negative samples \citep{Mikolov2013negative}, a window of size $10$ \citep{Levy2015improvingdist}, for $5$ iterations with initial learning rate set to the default $0.025$. 
GloVe is run with a window of size $10$, $x_{max} = 100$, $\beta=3/4$,
for $50$ iterations and initial learning rate of $0.05$ \citep{Pennington2014}.

In LexVec two window sampling alternatives are compared: $WS_{PPMI}$, which keeps the same fixed size $win = 2$ as used to create the $PPMI^*$ matrix; or $WS_{SGNS}$, which adopts identical SGNS settings ($win = 10$ with size randomization). 
We run LexVec for $5$ iterations over the training corpus.

All methods generate both word and context matrices ($W$ and $\tilde{W}$):   $W$ is used for SGNS, PPMI-SVD and $W + \tilde{W}$ for GloVe (following \cite{Levy2015improvingdist}, and $W$ and $W + \tilde{W}$ for LexVec.

For evaluation, we use standard word similarity and analogy tasks 
\citep{Mikolov2013negative,Levy2014linguisticregexplicit,Pennington2014,Levy2015improvingdist}. We examine, in particular, if LexVec weighted PPMI$^*$ factorization 
outperforms SVD, GloVe (weighted factorization of $\log\hat{M}$) and  Skip-gram (implicit factorization of the shifted PMI matrix), and compare
the stochastic and mini-batch approaches. 

Word similarity tasks are:\footnote{http://www.cs.cmu.edu/~mfaruqui/suite.html} WS-353 
Similarity (WSim) and Relatedness (WRel) 
\citep{Finkelstein2001},  MEN \citep{bruni2012distributional}, MTurk \citep{radinsky2011word}, RW \citep{Luong2013}, SimLex-999 \citep{hill2015simlex}, MC \citep{Miller1991}, RG \citep{Rubenstein1965}, and SCWS \citep{Huang2012}, calculated using cosine.  
Word analogy tasks are: Google semantic (GSem) and syntactic (GSyn) \citep{Mikolov2013sgoriginalnonegative} and MSR syntactic analogy dataset \citep{Mikolov2013linguisticregcontinuous}, 
using $3CosAdd$ and $3CosMul$ \citep{Levy2014linguisticregexplicit}.

\section{Results}
\label{sec:results}
Results for word similarity and for the analogy tasks are in \cref{tab:wordsim,tab:analogies}, respectively.
Compared with PPMI-SVD, LexVec performs better in all tasks.  
As they factorize the same $PPMI^*$ matrix, it is the loss weighting from window sampling that is an improvement over $L_2$ loss.
As expected, due to PPMI, LexVec performs better than SGNS in several word similarity tasks, but in addition it also does so on the semantic analogy task, nearly approaching GloVe. 
LexVec generally outperforms GloVe on word similarity tasks, possibly due to  the factorization of the PPMI matrix 
and to  
window sampling's weighting of negative co-occurrences. 

We believe LexVec fares well on semantic analogies because its vector-space does a good job of preserving semantics, as evidenced by its performance on word similarity tasks. We believe the poor syntactic performance is a result of the PPMI measure. PPMI-SVD also struggled with syntactic analogies more than any other task. \cite{Levy2015improvingdist} obtained similar results, and suggest that using positional contexts as done by \cite{Levy2014linguisticregexplicit} might help in recovering syntactic analogies.

In terms of configurations, WS$_{SGNS}$ performed marginally better than WS$_{PPMI}$. We hypothesize it is simply because of the additional computation. 
While $W$ and ($W+\tilde{W}$) are roughly equivalent on word similarity tasks,  $W$ is better for analogies. This is inline with  results for PPMI-SVD and SGNS models \citep{Levy2015improvingdist}.
Both mini-batch and stochastic approaches result in similar scores 
for all tasks. For the same parameter $k$ of negative samples, the mini-batch approach uses $2 * win_{WS_{PPMI}}$ times more negative samples than stochastic when using $WS_{PPMI}$, and $win_{WS_{SGNS}}$ times more samples when using $WS_{SGNS}$. Therefore, the stochastic approach is more computationally efficient while delivering similar performance.

\section{Conclusion and Future Work}
\label{sec:conclusions}
In this paper, we introduced LexVec, a method for low-rank, weighted factorization of the PPMI matrix that generates distributed word representations, favoring low reconstruction error on frequent co-occurrences, whilst accounting for negative co-occurrences as well. This is in contrast with PPMI-SVD, which does no weighting, and GloVe, which only considers positive co-occurrences. Finally, its PPMI factorization seems to better capture semantics when compared to the shifted PMI factorization of SGNS. 
As a result, it outperforms PPMI-SVD and SGNS in a variety of word similarity and semantic analogy tasks, and generally outperforms GloVe on similarity tasks. 

Future work will examine the use of positional contexts for  improving performance on syntactic analogy tasks. Moreover, 
we will explore further the hyper-parameter space 
to find globally optimal values for LexVec, and will experiment with the factorization of other matrices 
for developing alternative word representations.  

\section*{Acknowledgments}

This work has been partly funded by CAPES and by projects AIM-WEST (FAPERGS-INRIA 1706- 2551/13-7), CNPq 482520/2012-4, 312114/2015-0, ``Simplifica\c{c}\~ao Textual de Express\~{o}es Complexas", sponsored by Samsung Eletr\^{o}nica da Amaz\^{o}nia Ltda. under the terms of Brazilian federal law No. 8.248/91.

\bibliography{serveboy_manual,confs}
\begin{footnotesize}
\bibliographystyle{acl_natbib}
\end{footnotesize}
\appendix

\end{document}